\documentclass[journal]{IEEEtran}

\usepackage{cite}
\usepackage{graphicx}
\usepackage{subcaption}
\usepackage{multirow}
\usepackage{amsmath}
\usepackage{array}
\usepackage{url}

\begin{document}

\title{A Novel Approach to Curiosity and Explainable Reinforcement Learning via Interpretable Sub-Goals} 

%
%
%

\author{
		Connor~van~Rossum,~
		Candice~Feinberg,
		Adam~Abu~Shumays,
		Kyle~Baxter, 
		Benedek~Bartha 
		
\thanks{Manuscript received April 13, 2021. (\emph{All authors contributed equally to this work})}
\thanks{The authors are with BrightMind, Bright Media Inc., California USA (email: connor@brightagi.com, 
	candice@brightagi.com, adam@brightagi.com, kylebaxter@brightagi.com, benedek.bartha@wolfson.ox.ac.uk).}
\thanks{Digital Object Identifier ....}}

%



\maketitle

\begin{abstract}
Two key challenges within Reinforcement Learning involve improving (a) agent learning within environments with sparse extrinsic rewards and (b) the explainability of agent actions. We describe a curious subgoal focused agent to address both these challenges. We use a novel method for curiosity produced from a Generative Adversarial Network (GAN) based model of environment transitions that is robust to stochastic environment transitions. Additionally, we use a subgoal generating network to guide navigation. The explainability of the agent's behavior is increased by decomposing complex tasks into a sequence of interpretable subgoals that do not require any manual design. We show that this method also enables the agent to solve challenging procedurally-generated tasks that contain stochastic transitions above other state-of-the-art methods.
\end{abstract}


%

\section{Introduction}
%
%
%
%
\IEEEPARstart{D}{eep} Reinforcement Learning (DRL) has been tremendously successful. Agents have been trained to defeat a champion Go player \cite{alphago}, outperform humans in Atari games \cite{muzero, 57} and to perform a variety of difficult robotic tasks \cite{robobench, gSDE}. Two of the major challenges in DRL are the problem of sparse rewards and the lack of explanablity provided by the algorithms. Random exploration in environments with extremely sparse rewards, which is the core exploration method in many RL approaches, is unlikely to receive a sufficient reward signal to successfully train the agent \cite{rewards}. The Deep Neural Networks (DNNs) utilized by DRL also lack explainablity of their decision making. Several works have focused on investigating methods for local interpretability, most notably attribution methods, primarily in supervised learning \cite{lime, shapley, grad}. These methods generally show which inputs were most influential in producing the DNN's output as per the algorithms criteria. Similar methods have been utilized in DRL in addition to reward decomposition \cite{rewarddecomp} and action influence \cite{clens}.

To improve upon the tools available to increase exploration and explainability with DRL, we present a novel method for goal orientated environments in which the agent learns to propose subgoals. The agent is composed of a subgoal generating network, a navigator network that learns to move towards the current subgoal, and a Generative Adversarial Network (GAN) tasked with next state generation. The GAN serves to build an understanding of object interaction and the effects of actions within the environment. Subgoals are set relative to the observation space, where a goal value for one element of the space is selected by the subgoal generator. The navigator network takes actions within the environment with the intent of fulfilling the chosen subgoal. The decomposition of a complex task into a sequence of simpler subgoals increases the human interpretability of the agent's policies and decisions. Exploration efficiency and performance is also increased because of the additional feedback the subgoals provide. 

Many authors have pointed out that agents that quantify novelty through prediction errors tend to get attracted to transitions that are stochastic in nature \cite{curiosity, skill, curbench} decreasing the effectiveness of curiosity driven exploration. We address this by implementing a modification of BicycleGAN \cite{bgan} as our prediction network. BicycleGAN is a multimodal image generator, able to generate a variety of possible images from the same input. With BicycleGAN, our agent predicts possible next states and determines impossible transitions removing the need to make a single deterministic prediction. 

We evaluate our model on procedurally generated environments, which are naturally challenging as agents need to deal with a parameterized family of tasks, resulting in large observation spaces and making memorizing trajectories infeasible. It is in these environments that the agent must learn to generalize its policies.

\section{Related Work}
Intrinsic motivation has been a popular method to improve the exploration abilities of agents within environments yielding sparse rewards. Novelty and curiosity are two of the most prominent sources of intrinsic motivation. Where novelty is primarily driven by reaching unfamiliar states, curiosity seeks to better understand or predict environment dynamics. One of the methods to power curiosity is next state prediction, where future states that have greater prediction errors receive greater rewards to encourage the agent to take actions within states that it is not yet able to accurately predict the result of. 

Multi-Goal Hierarchical Reinforcement learning (MGHL) \cite{subgoals} by Xing utilizes a manager network, which sets auxiliary control tasks as subgoals that direct a worker network. The manager primarily sets subgoals where the worker is rewarded for either changing the selected pixel blocks or changing high level features where the magnitude of the reward is proportional to the magnitude of the change irrespective of the direction. We contrast this to our work where our agent generates subgoals by selecting one observation element and one particular value where the resulting subgoal is to change the chosen observation element to the chosen value. 

Another similar sub-task orientated work is the Modular Multitask Reinforcement Learning with Policy Sketches \cite{subtask}. The approach uses a high level policy to select sub-tasks that a worker network learns to carry out. In contrast to our work, the high level policy requires hand crafted sub tasks for each unique environment and does not include any predictive models of the environment. 

Adversarially Motivated Intrinsic Goals (AMIGo) \cite{amigo} sees a teacher network generate goal positions for a student network to learn to reach. The teacher network is rewarded by generating goals that the student is able to reach; but not for those reached too quickly. In contrast to our model which only uses the agent's partial view, the teacher network sets goals relative to the full observation space. Additionally, the primary purpose of the teacher network is to drive exploration with a single goal increasing in difficulty rather than increase the explainability of the agent through a sequence of subgoals.

\section{Curious Sub-Goal focused Agent}

Our Curious Sub-Goal focused agent (CSG) is composed of three subsystems: a next state generator powered by a GAN, the agent navigation network, and a subgoal generating network. 

\subsection{Next State Generation}
Previous works on next state prediction primarily focus on producing a deterministic prediction and comparing this prediction to the actual next state to produce a loss function, with an optimizer using the loss function to improve future predictions. This method of prediction does poorly in environments with a lot of noise present or if there are stochastic elements present. By using a GAN to generate the next state, the network is not predicting the next state but instead producing one possible next state. In taking this approach the agent is able to learn which transitions are possible and which are impossible, and in doing so gain robustness to noise and stochasticity. 

In order to be able to predict a variety of possible transitions we learn a low-dimensional latent space $\mathbf{z} \in \mathbf{R}^\mathbf{Z}$ which encapsulates the stochastic aspects of the transition. We then learn a deterministic mapping $G(\mathbf{s}_t, a_t, \mathbf{z}_t) \to \mathbf{\hat s}_{t+1}$. Where $a_t$ is the action at time step $t$, $\mathbf{s}_t$ is the state at time step $t$, and $\mathbf{\hat s}_{t+1}$ is a generated configuration of what the state could be at time step $t+1$. We draw the latent vector $\mathbf{z}_t$ from a prior distribution $p(\mathbf{z}_t)$; in this work we use a standard Gaussian distribution $\mathcal{N}(0,1)$. A discriminator network, $D$, is trained to detect synthetic transitions generated by $G$: $D(\mathbf{s}_t, a_t,\mathbf{u}_{t+1}) \to d_{t}$, where $\mathbf{u}_{t+1} \in \{\mathbf{s}_{t+1}, \mathbf{\hat s}_{t+1}\}$, and $d_t \in (0,1)$ corresponds to the probability that $\mathbf{u}_{t+1}$ is synthetic ($\mathbf{u}_{t+1} = \mathbf{\hat s}_{t+1}$).

We use the generator $G$ and discriminator $D$ as a source for an intrinsic curiosity reward similar to those produced from next state prediction error. In order to produce in informative curiosity reward, we force the generator $G$ to attempt to generate a prediction that matches the true next state $\mathbf{s}_{t+1}$. We directly map $\mathbf{s}_{t+1}$ to the latent space $\mathbf{z}$ using an encoding function, $E(\mathbf{s}_{t+1}) \to \mathbf{z}_t^\text{enc}$. The generator $G$ then uses a sample of the encoded latent space $\mathbf{z}_t^\text{enc}$, the state $\mathbf{s}_{t}$ and action $a_t$ to synthesize the desired output $\mathbf{\hat s}_{t+1}^\text{enc}$. This process can be likened to the reconstruction of $\mathbf{s}_{t+1}$ using an auto-encoder. We define the reconstruction error of $\mathbf{s}_{t+1}$ to be $|| \mathbf{s}_{t+1} - \mathbf{ \hat s}_{t+1}^\text{enc} ||$, which is used as a component of the curiosity reward function. 

We define the curiosity based reward utilizing the reconstruction error and the discriminator's output $d_t$ predicting a synthetic transition:

\begin{equation}
	r_t^c = \alpha || \mathbf{s}_{t+1} - \mathbf{ \hat s}_{t+1}^\text{enc} || d_t
\end{equation}

where $\alpha \in [0,\infty)$ is a scaling hyper parameter.

\subsection{Navigation Network}
We consider the traditional RL framework of a Markov Decision Process with a state space $S$, a set of actions $A$ and a transition function $p(\mathbf{s}_{t+1}|\mathbf{s}_t, a_t)$ which specifies the distribution over next states given a current state and action. At each time-step $t$, the agent in state $\mathbf{s}_t \in S$ takes an action $a_t \in A$  by sampling from a goal-conditioned stochastic policy $\pi(a_t|\mathbf{s}_t,g_t;\theta_\pi)$ represented as a neural network with parameters $\theta_\pi$ where $g_t$ is the goal. We assume that some goal verification function $v$ can be specified such that $v(\mathbf{s}_t,g_t) = 1$, if the state $\mathbf{s}_t$ satisfies the goal $g_t$, and 0 otherwise. The undiscounted intrinsic goal reward $r^{g_t}$ is defined to be:

\begin{equation}
	r^{g_t}_t =
	\begin{cases}
		r & \text{if } v(\mathbf{s}_t, g_t) = 1\\
		0 & \text{otherwise}
	\end{cases}       
\end{equation}

where $r>0$  is a hyper-parameter. The navigation network is trained to maximize the discounted expected reward, $R^{Nav}_t$:

\begin{equation}
	R^{Nav}_t = \mathbf{E}[\sum_{k=0}^{T}\gamma^k (r^{g_t}_{t+k} + r^{c}_{t+k})]
\end{equation}
where $\gamma \in [0,1)$ is the discount factor and $r^c$ is the curiosity reward.

\subsection{Sub-Goal Generation}
The subgoal generator neural network is defined as $SG(\mathbf{s}_t,g_0) \to g_t$, where $g_t$ is the current goal, and $g_0$ is the initial goal set by the environment. Possible goals are set relative to the agent's observation space. It selects an input position and a goal value for that position. The navigator network is then tasked with altering the selected input to be the goal value. A new subgoal can be proposed at the start of an episode, once the previous one is reached or the previous one is abandoned. The agent abandons subgoals if it exceeds a hyper-parameter that sets the step limit to attempt to reach the goal. Finally, the network can also choose to set no subgoal, which is taken when the network selects the current goal as the subgoal.

The subgoal generator network is trained to maximize the discounted expected reward:

\begin{equation}
	R^{SG}_t = \mathbf{E}[\sum_{k=0}^{T}\gamma^k (r^{c}_{t+k} + \frac{v(\mathbf{s}_{t+k}, g_{t+k}) + \beta}{1+ \beta} r^{e}_{t+k})]
\end{equation}

where $\gamma \in [0,1)$ is the discount factor, $r^c$ is the curiosity reward, $r^e$ is the extrinsic reward within the environment, $v(\mathbf{s}_t, g_t)$ is the goal verification function, and $\beta \in [0,\infty)$ is a scaling parameter which determines the penalty applied to the extrinsic reward when the goal does not align with the reward.

\section{Experiments}
We use a modified version of the TorchBeast implementation of IMPALA \cite{impala} (\url{https://github.com/facebookresearch/torchbeast}) as the framework for our code.

We establish our proof of concept within the challenging and scaleable environment of minigrid (\url{https://github.com/maximecb/gym-minigrid}). This provides a good test bed for the agent as the observations are symbolic rather than high dimensional allowing for clear and interpretable subgoals.

\subsection{Environments}
\begin{figure}
	\centering
	\begin{subfigure}{.24\textwidth}
		\centering

		\includegraphics[width=.8\linewidth]{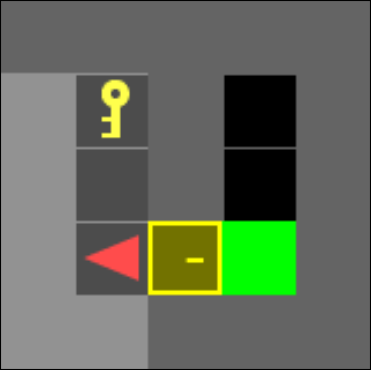}
		\caption{$5 \times 5$ Door \& Key}		
		\label{dk5}
		\vspace{4mm}
	\end{subfigure}%
	\begin{subfigure}{.24\textwidth}
		\centering

		\includegraphics[width=.8\linewidth]{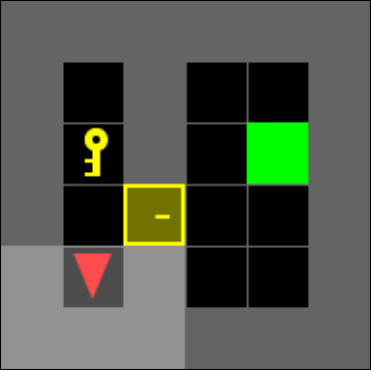}
		\caption{$6 \times 6$ Door \& Key}		
		\label{dk6}
		\vspace{4mm}
	\end{subfigure}
	
	\begin{subfigure}{.24\textwidth}
		\centering

		\includegraphics[width=.8\linewidth]{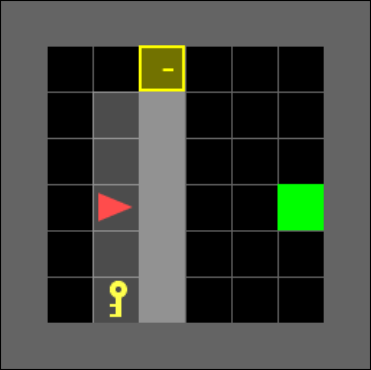}
		\caption{$8 \times 8$ Door \& Key}
		\label{dk8}
	\end{subfigure}
	\begin{subfigure}{.24\textwidth}
		\centering

		\includegraphics[width=.8\linewidth]{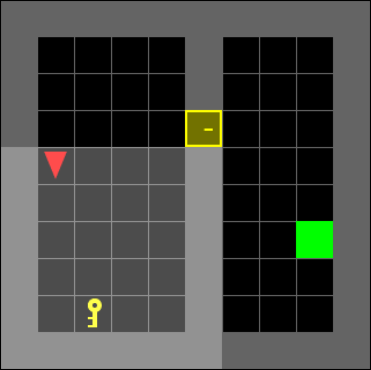}
		\caption{$10 \times 10$ Door \& Key}
		\label{dk10}
	\end{subfigure}
	\vspace{1mm}
	\caption{Examples of Door \& Key environments. To complete the environment the agent (red triangle) must proceed through the locked door (yellow box) to the green goal zone in the adjacent room. Vision is only in the direction the agent faces (red triangle) and expands from a $5 \times 5$ box in (\protect\subref{dk5}, \protect\subref{dk6}, \protect\subref{dk8}) to a $9 \times 9$ box in (\protect\subref{dk10}). Objects within the agent's vision are indicated by a lighter gray shading.}
	\label{dk_ex}
\end{figure}

We use a modified version of Door \& Key environments within the minigrid package (Figure \ref{dk_ex}). The environment randomizes the position of the dividing wall, door, and key, as well as the agent's starting position and orientation. We alter the environment to also randomize the location of the agent's target position within the second room. The randomization adds increased difficulty by increasing the number of possible trajectories and adds stochastic transitions into the environment.

The environments are comprised of $N \times N$ tiles, where each tile contains at most one of the following objects: door, key, wall, goal. If it contains none, it is the `empty' object. The agent has limited vision, seeing within a $M \times M$ square with the agent positioned on a center edge of the square. The agent cannot see behind walls or closed doors even if they fall within the vision range. Such objects are assigned as `unseen' rather than their true values. All objects and their attributes are transformed using an embedding layer unique for each network before being fed into the remainder of the network.  The navigator is able to perform the following actions: turn left, turn right, move forward, pick up an object, drop an object, or toggle (interact with objects such as opening a door). 

\subsection{Implementations}
The navigator and subgoal generator have a network architecture consisting of a fully connected layer with rectified linear units, two LSTM layers, followed by a final fully connected layer. They are trained using the TorchBeast \cite{torchbeast} implementation of IMPALA \cite{impala}. 

The GAN's encoder simply consists of three fully connected layers interleaved with a hyperbolic tangent activation function. Similarly, both the generator and discriminator consist of four fully connected layers interleaved with a hyperbolic tangent activation function. In addition, the singular output from the final layer of the discriminator is activated by a sigmoid function. We train the GAN using the same methods as BicycleGAN \cite{bgan}, with one exception. As BicylceGAN is designed for continuous data sets, we replace the standard generator loss function with the MaliGAN \cite{mali} gradient estimator. 

For comparison, we implement 3 alternate methods. We use IMPALA without any intrinsic motivation as a standard deep RL baseline. Additionally, we include two alternative methods utilizing intrinsic motivation. Random Network Distillation Exploration (RND) \cite{rnd} which uses next state feature prediction error as a curiosity based form of intrinsic motivation, where features are produced from a random neural network. AMIGO is also trained where the teacher and student network architectures match those described in the AMIGO paper \cite{amigo} and accompanying code \url{https://github.com/facebookresearch/adversarially-motivated-intrinsic-goals}. The AMIGO networks receive the full absolute view of the environment, as opposed to the agent centric partial view the other methods receive. 
Both RND and IMPALA share the same network architecture as our navigator and subgoal generator networks. All three alternate methods are trained using the TorchBeast framework.

\begin{table*}[t]
	\centering
	\setlength{\extrarowheight}{2pt}
	\begin{tabular}{p{0.17\textwidth}>{\centering}p{0.17\textwidth}>{\centering}p{0.17\textwidth}>{\centering\arraybackslash}p{0.17\textwidth}>{\centering\arraybackslash}p{0.17\textwidth}}
		\hline
		\multirow{2}{*}{Model}&\multicolumn{3}{c}{Door \& Key Mean Extrinsic Reward ($N \times N$ size)}\\\cline{2-5}
		& $5 \times 5$ & $6 \times 6$ & $8 \times 8$ & $10 \times 10$ \\
		\hline
		CSG & $0.85 \pm 0.01$  & $0.89 \pm 0.00$ & $0.88 \pm 0.01$ & $0.92 \pm 0.01$ \\
		IMPALA & $0.91 \pm 0.00$ & $0.90 \pm 0.00$ & $0.94 \pm 0.01$ & $0.00 \pm 0.00$ \\
		RND & $0.87 \pm 0.01$ & $0.86 \pm 0.00$ & $0.89 \pm 0.00$ & $0.82 \pm 0.01$ \\
		AMIGO & $0.88 \pm 0.01$ & $0.86 \pm 0.00$ & $0.87 \pm 0.02$ & $0.81 \pm 0.02$ \\
		\hline
	\end{tabular}
	\caption{Comparison of mean extrinsic reward at the end of training. $5\times 5$ was run for 5,000,000 steps, $6 \times 6$ for 30,000,000 steps, $8 \times 8$ for 60,000,000 steps, and $10 \times 10$ for 90,000,000 steps.}
	\label{dkres}
\end{table*}

\subsection{Results and Discussion}
We use the mean extrinsic reward at the end of training to evaluate the performance of each agent. Where the extrinsic reward defined in the environment is $r^e_t = 1- (0.9t)/t_\text{max}$. We summarize our results in Table \ref{dkres}.

IMPALA is unable to reach the goal once the environment space becomes too large given the lack of intrinsic motivation for exploration. However it performs better in the low dimensional spaces due to the lack of reward noise the other methods utilize. RND and AMIGO perform comparatively across all environment dimensions. Our agent, CSG, performs better than both RND and AMIGO in the highest dimensional environment.

\begin{figure}
	\centering
	\captionsetup[subfigure]{justification=centering}
	\begin{subfigure}{.24\textwidth}
		\centering  
		\includegraphics[width=.8\linewidth]{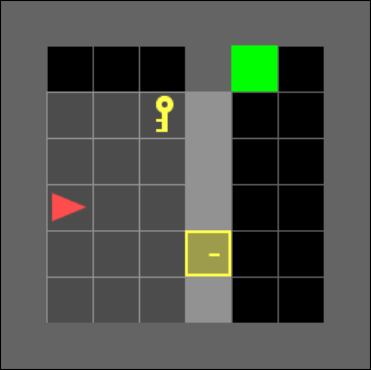}
		\caption{ Time step $0$. \protect\\ Subgoal: Pick up key}
		\label{goalfig_1}
		\vspace{4mm}
	\end{subfigure}
	\begin{subfigure}{.24\textwidth}
		\centering
		\includegraphics[width=.8\linewidth]{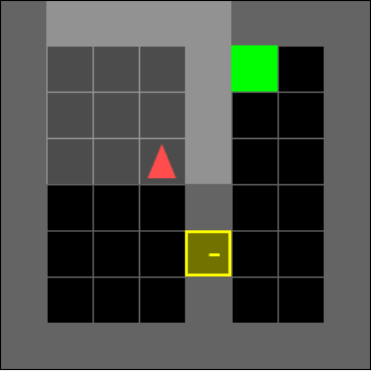}
		\caption{Time step $5$. \protect\\ New subgoal: Go to door.}
		\label{goalfig_2}
		\vspace{4mm}
	\end{subfigure} \\
	\begin{subfigure}{.24\textwidth}
		\centering
		\includegraphics[width=.8\linewidth]{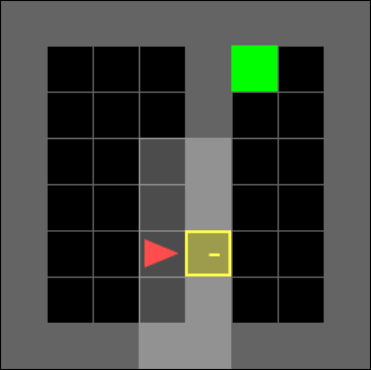}
		\caption{Time step $10$. \protect\\ New subgoal: Go to goal.}
		\label{goalfig_3}
	\end{subfigure}
	\begin{subfigure}{.24\textwidth}
		\centering
		\includegraphics[width=.8\linewidth]{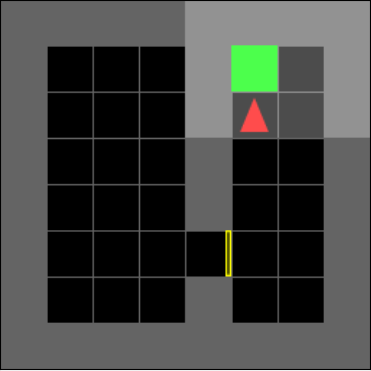}
		\caption{Time step $17$. \protect\\ Final step before episode ends}
		\label{goalfig_4}
	\end{subfigure}
	\vspace{1mm}
	\caption{Trained Agent completing the $8 \times 8$ environment. Frames (a) - (c) show a new subgoal generation, in addition (d) shows the last frame before the episode ends.}
	\label{goalfig}
\end{figure}

The novel subgoal generation has the added benefit of increasing the agent's explainablity. Figure \ref{goalfig} highlights an example set of goals used to direct the navigator network in order to solve the environment. The subgoals listed are the human interpretation of the subgoal generator output. In figure \ref{goalfig} the agent receives a $5 \times 5$ field of view:

\begin{equation*}
	\begin{bmatrix}
		\mathbf{b}_{11} & \mathbf{b}_{12} & \mathbf{b}_{13} & \mathbf{b}_{14} & \mathbf{b}_{15} \\
		\mathbf{b}_{21} & \mathbf{b}_{22} & \mathbf{b}_{23} & \mathbf{b}_{24} & \mathbf{b}_{25} \\
		\mathbf{b}_{31} & \mathbf{b}_{32} & \mathbf{b}_{33} & \mathbf{b}_{34} & \mathbf{b}_{35} \\
		\mathbf{b}_{41} & \mathbf{b}_{42} & \mathbf{b}_{43} & \mathbf{b}_{44} & \mathbf{b}_{45} \\
		\mathbf{b}_{51} & \mathbf{b}_{52} & \mathbf{b}_{53} & \mathbf{b}_{54} & \mathbf{b}_{55} \\
	\end{bmatrix}
\end{equation*}

where $\mathbf{b}_{ij}$ is an object. The field of view moves with the agent such that the agent is always positioned at $\mathbf{b}_{35}$, with $\mathbf{b}_{34}$ directly in front. At time step $0$, in Figure \ref{goalfig_1}, the subgoal generator selects $\mathbf{b}_{35}$ to be changed to the object index value corresponding to the yellow key, which can be interpreted as the agent holding the key, and hence the subgoal is to pick up the key. Similarly, at time step $5$, in Figure \ref{goalfig_2}, the subgoal generator selects $\mathbf{b}_{34}$ to be changed to the object index value corresponding to the locked yellow door, which we interpret as the agent setting going to the door as its subgoal. Finally, at time step $10$, in Figure \ref{goalfig_3}, $\mathbf{b}_{35}$ is selected to be changed to the object index value corresponding to the green goal zone as its subgoal, which is also the overarching goal set by the environment. 

The ability to generate explicit subgoals has two benefits: the first is to developers, as it gives a window into the training process, showing stages along the way to the agent’s final functionality, that are accessible to human language, logic and reasoning. Therefore, more interventions are possible, giving the ability to craft and debug the training process at a finer grain than conventional methods. The second benefit is to end users of the agent. These users will be able to understand in human terms how and why the agent made the decisions it did, leading to greater trust in the agent. Along this track there is a future long-term benefit to end users in that their feedback can be integrated as reward to train the agent, giving them the ability to teach the agent according to their preferences. Users will get a greater measure of control and insight into the agent's reasoning and decision criteria.

\section{Conclusion}
In this work we propose a method for developing a curious subgoal focused agent that has a robust method for driving curiosity and that decomposes complex tasks using interpretable subgoals increasing both exploration and explainability. In future work we would like to expand the testing environments to cement CSG as a new state of the art hard exploration method.

In this implementation of the CSG, applications are limited. They are restricted to discrete observation spaces with basic dynamics. However, it may be easily incorporated to run with any modern DRL agent operating within these constraints as it needs no hand-crafted features or demonstrations. In future work we also would like to include abstract concepts and representations as possible subgoals. For example if the pixel input version of gridworld were used, we would like to enable concepts for `key' and `door' created by the agent to be used rather than any specific pixel.


%




\ifCLASSOPTIONcaptionsoff
  \newpage
\fi

\bibliographystyle{IEEEtran}

\bibliography{CSG}

\end{document}